\RequirePackage{fix-cm}

\documentclass[14pt,twocolumn,natbib,headings=normal,enabledeprecatedfontcommands]{svjour3}          
 
\usepackage{graphicx}
\usepackage{float}
\usepackage[draft]{hyperref}
\usepackage[utf8x]{inputenc}			
\usepackage[T1]{fontenc}				
\usepackage[english]{babel}				
\usepackage{microtype}					
\usepackage{amsmath}					
\usepackage{graphicx}					
\usepackage{blindtext}					
\usepackage{booktabs}					
\usepackage{color}						
\usepackage{nomencl}					
\usepackage{caption}					
\usepackage{subfig}					
\expandafter\def\csname 
ver@subfig.sty\endcsname{} 			
\usepackage{hyperref}					
\usepackage{blindtext}					
\usepackage{amsmath}
\usepackage{natbib}

\usepackage{bm}	
\usepackage{color}										
\usepackage{colortbl}
\usepackage{here} 
\definecolor{dunkelgrau}{rgb}{0.8,0.8,0.8}
\definecolor{hellgrau}{rgb}{0.95,0.95,0.95}
\usepackage{epstopdf}
\usepackage{multirow}
\usepackage{hyperref}

\usepackage{graphicx}

\usepackage{tikz}
\usetikzlibrary{shapes,arrows}

\usepackage{placeins}
\usepackage{tabularx}
\usepackage{enumitem}

\usepackage{upgreek}
\usepackage{capt-of}

\usepackage{xcolor}

\usepackage{siunitx}
\usepackage{scrextend}
\addtokomafont{labelinglabel}{\sffamily}

\usepackage{epstopdf}
\AppendGraphicsExtensions{.tif}

\usepackage{amssymb}
\usepackage{amsmath}

\usepackage[T1]{fontenc}
\usepackage{lmodern}

\usepackage{soul,color}
\newcommand{\tenso}[2][1]{
	\ifnum#1=1 
	\vec{\boldsymbol{#2}}
	\else
	\ifnum#1=2  
	\boldsymbol{#2}
	\else
	\ifnum #1=3 
	\boldsymbol{\mathfrak{\unbold{#2}}}
	\else
	\ifnum #1=4 
	\boldsymbol{\mathcal{#2}}
	\else
	\ifnum #1=0 
	\unbold{#2}
	\fi
	\fi
	\fi
	\fi
	\fi
}

\usepackage{mdframed}
\usepackage{lipsum}

\newmdtheoremenv{theo}{Definition}

\usepackage{multirow}

\usepackage[affil-it]{authblk}

\begin{document}

\title{Automated Quality Control of Vacuum Insulated Glazing by Convolutional Neural Network Image Classification}


\author{Henrik Riedel* \and Sleheddine Mokdad \and Isabell Schulz \and Cenk Kocer \and Philipp Rosendahl \and Jens Schneider \and Michael A. Kraus \and Michael Drass}

\institute{
	*corresponding author: M.Sc. Henrik Riedel \\ 
	M.Sc. Henrik Riedel \\
	TU Darmstadt - Institute for Structural Mechanics and Design\\
	\email {riedel@ismd.tu-darmstadt.de}\\
	+49 6151 1623011\\
	Preprint. Submitted for review.
}

\maketitle

\begin{abstract}
Vacuum Insulated Glazing (VIG) is a highly thermally insulating window technology, which boasts an extremely thin profile and lower weight as compared to gas-filled insulated glazing units of equivalent performance. The VIG is a double-pane configuration with a submillimeter vacuum gap between the panes and therefore under constant atmospheric pressure over their service life. Small pillars are positioned between the panes to maintain the gap, which can damage the glass reducing the lifetime of the VIG unit. To efficiently assess any surface damage on the glass, an automated damage detection system is highly desirable. For the purpose of classifying the damage, we have developed, trained, and tested a deep learning computer vision system using convolutional neural networks. The classification model flawlessly classified the test dataset with an area under the curve (AUC) for the receiver operating characteristic (ROC) of 100\%. We have automatically cropped the images down to their relevant information by using Faster-RCNN to locate the position of the pillars. We employ the state-of-the-art methods Grad-CAM and Score-CAM of explainable Artificial Intelligence (XAI) to provide an understanding of the internal mechanisms and were able to show that our classifier outperforms ResNet50V2 for identification of crack locations and geometry. The proposed methods can therefore be used to detect systematic defects even without large amounts of training data. Further analyses of our model’s predictive capabilities demonstrates its superiority over state-of-the-art models (ResNet50V2, ResNet101V2 and ResNet152V2) in terms of convergence speed, accuracy, precision at 100\% recall and AUC for ROC.

\end{abstract}

\section{Introduction}
As this work connects the domains of deep learning and glass structures, it is crucial for comprehension of this paper to establish an understanding of the assembly of VIGs, the nature and occurence of their typical cone-shaped cracks, as well as applicable deep learning methods for automated damage evaluation of VIGs.

\subsection{Vacuum Insulated Glazing}
\begin{figure}[t]
	\centering
    \includegraphics[width=\linewidth]{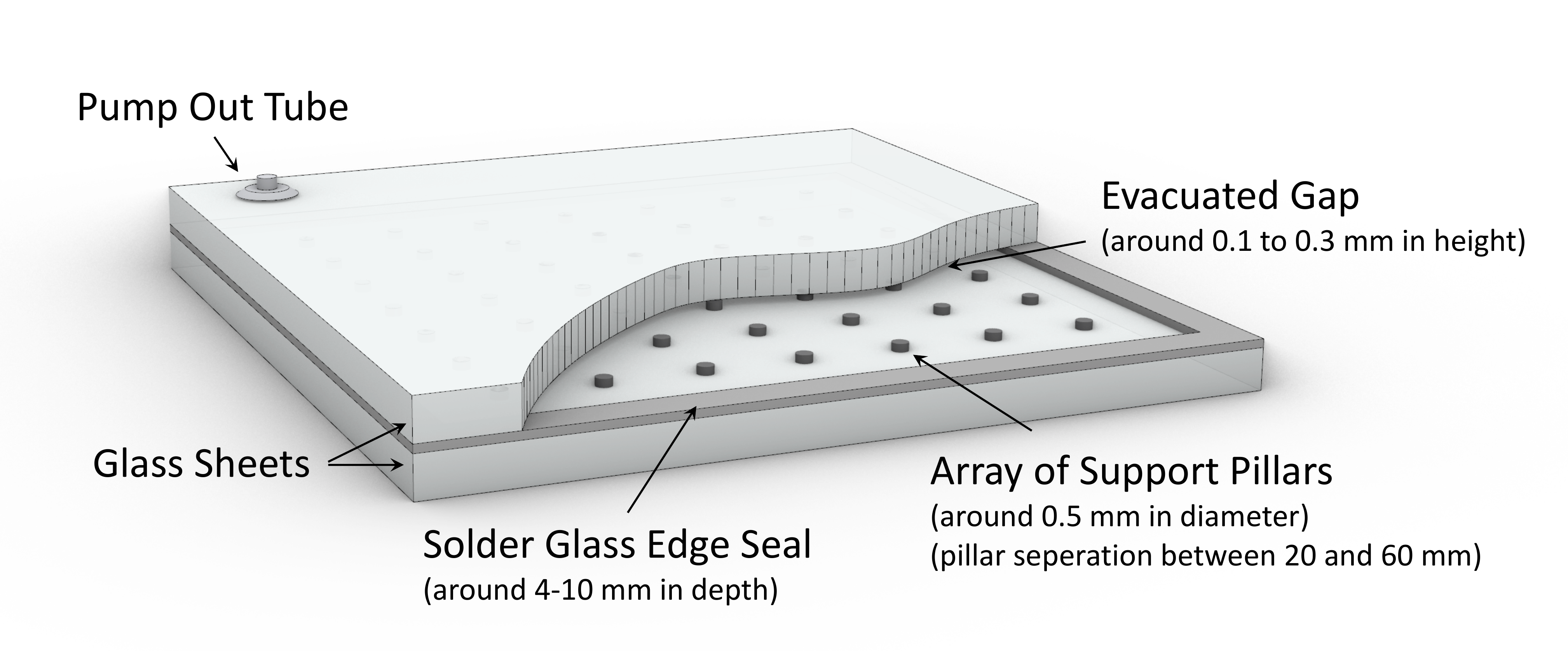}
	\caption{Basic components of a VIG.}
	\label{fig:vig components}
\end{figure}

\begin{figure}
	\centering
    \includegraphics[width=\linewidth]{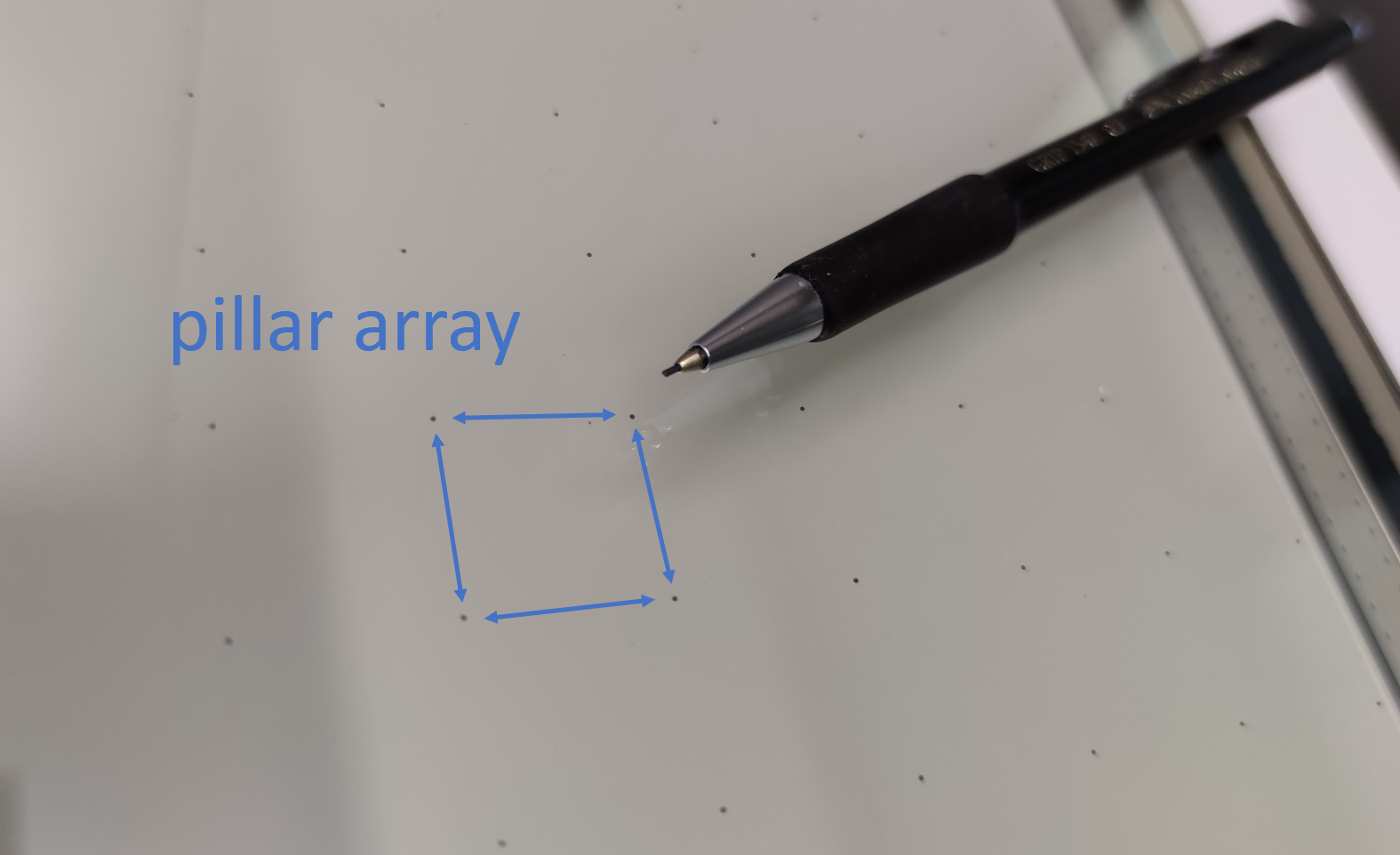}
	\caption{Picture of a VIG sample with pencil for scale. Red arrows show the distance between 4 different pillars to highlight their pattern.}
	\label{fig:vig real}
\end{figure}

Vacuum Insulated Glazing (VIG) units are highly thermally insulating assemblies similar in design to the traditional Insulated Glazing Units (IGU) \citep{collins1991design,collins1999vacuum}; see fig.~\ref{fig:vig components} and \ref{fig:vig real}. Unlike IGUs, the thermal conductance (U-value) of VIGs does not rely on a low-conductivity gas. VIGs are constructed from two panes of glass separated by an evacuated submillimeter gap and, unlike the IGU, the performance of the VIG is independent of the gap size. The required level of vacuum in the gap is ensured by hermetically sealing the perimeter of the two glass panes, where typically solder glass (a glass frit material) is used. To maintain the separation of the glass panes under atmospheric pressure, a regular square array of disk-shaped spacers (also called pillars) is placed between the glass panes (see fig.~\ref{fig:vig real}). The typical construction of a VIG unit is illustrated in fig.~\ref{fig:vig components}.

At far distance from the edge of the glass, the heat flow through VIGs is primarily conduction of heat through the pillar array and surface-to-surface radiation between the glass panes. The vacuum level in the gap, that is the residual gas pressure, is chosen such that gaseous conduction is negligible. Considering the design landscape of VIGs, the U-value of VIGs can be in the range of about 1 to 0.3 \si{W.m^{-2}.K^{-1}} \citep{lee2018new, simko1999determination}. Ultimately, VIGs are window elements that are thin and light weight, with a potential U-value limit as low, if not lower, than existing triple pane IGUs, which are much thicker and heavier. When designing for a low U-value, the impact of design choices on the mechanical strength of the VIG must be carefully considered. The most critical design impact is the pillar shape, size, and the array spacing (distance between pillars).

\begin{figure}
	\centering
	\includegraphics[width=\linewidth]{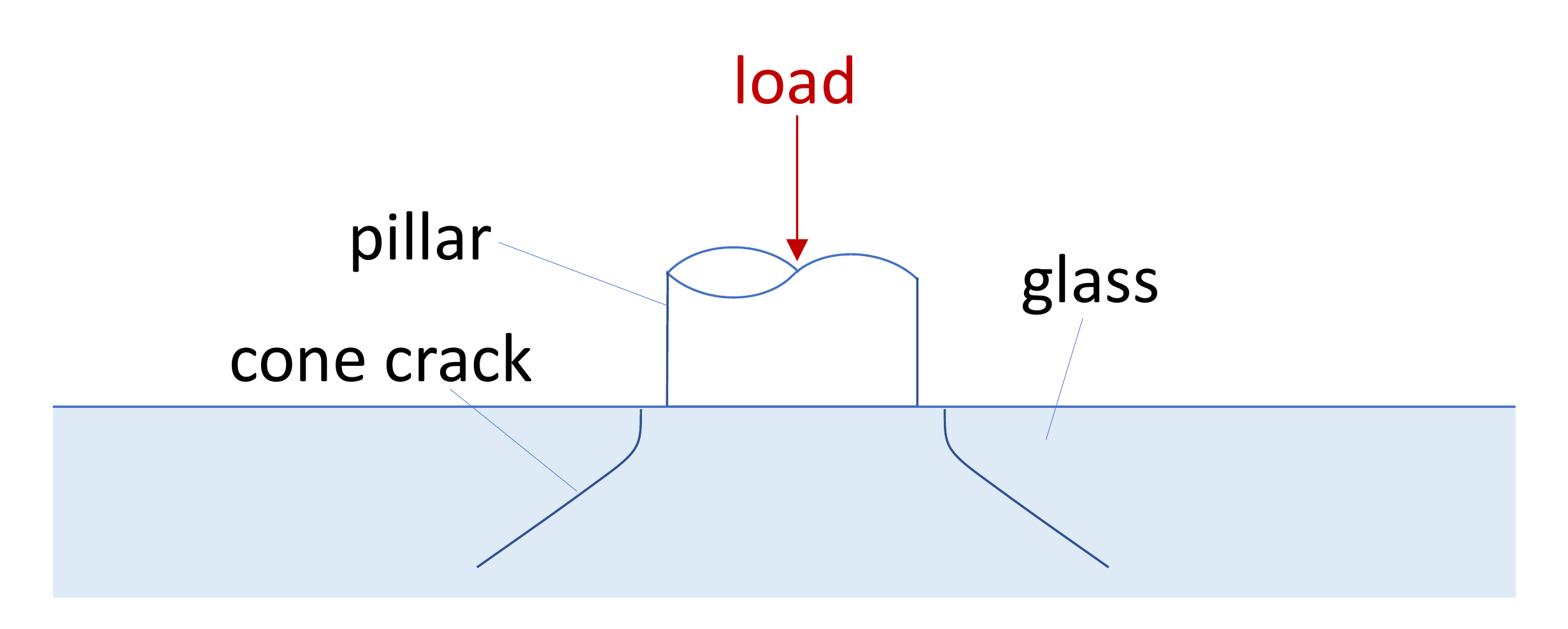}
	\caption{An illustration of the pillar contact on the glass surface, and the typical cone crack}
	\label{fig:conecrack}
\end{figure}

Typically, pillars are made from a stainless steel alloy, and they are about 0.5 mm in diameter, 0.2 mm in height, and are positioned between the glass panes as a regular array with a spacing of about 20 mm; this results in a unit U-value of about \si{1.W.m^{-2}.K^{-1}} depending on the low emissivity coating used \citep{collins1991design}. In order to reduce further the U-value, either the pillar size can be decreased or the array spacing can be increased; in most cases the best performing low emissivity coatings are used. In both these options the consequences are, (1) increased stress concentrations due to the higher level of contact pressure on the glass and, (2) larger deflections of the unsupported glass between the pillars. or design and production purposes, the stress field in the glass due to pillar contact is of great interest. Unlike in other areas, the contact stress field is highly localized and the stress gradient is high. This situation corresponds to the well-known indentation configuration (blunt indentation) and produces stable Hertzian cone cracks as shown in fig.~\ref{fig:conecrack}, which themselves do not cause catastrophic failure of the VIG unit without excessive increases of external loads \citep{Hertz1882,Mouginot1985,kocer2003}. Simply, in most external load cases of a VIG, pillar-induced cone cracks, which do not casue instantaneous failure of the pane, can be considered as an equivalent increase in the surface flaw statistics of the underlying glass pane; that is, the cone cracks are potential failure origins in future load scenarios.

As with any glass product, the final design choice is highly influenced by the residual strength of the underlying components, which in the case of the glass panes, is defined by the surface flaw statistics \citep{meyland2021tensile}. Even if a conservative safety factor is used, the fact that in a VIG contact cone cracks can be produced during manufacturing or transport, there is a significant uncertainty in the actual strength of the glass elements \citep{fischer1995architectural}. This is of particular concern since the service lifetime of a VIG unit should be beyond 30 years, where the unit must sustain constant atmospheric pressure, and the cumulative impact of thermal, wind, and even debris impact loads. It is further possible that the cracks develop due to nonuniformity issues in the manufacturing of the pillars and the process used to place them on the glass surface. In a \si{1.m^2} VIG unit, depending on the design, there can be anywhere between 620 to 2,500 pillars \citep{collins1991design}.   

To increase the overall durability of the VIG product, it would be favourable to visually examine VIG units during post-production, to identify cone cracks, determine the structural implications, and remove those units that would potentially fail during service. Unfortunately, to the best of our knowledge, currently there is no visual inspection, either by eye or machine scanning, that would be sensitive enough to identify surface cone cracks with high precision at an efficient speed. Using a high-resolution camera with magnification, a human operator examining all pillars in a unit would be time consuming and laborious, adding greatly to production times and cost. Hence, an automated process, not only to undertake high-speed imaging but also to perform high speed image analysis, is highly desirable. The know-how and hardware for high resolution and high-speed processes are readily available, where scan times could be as low as 5-10 sec per unit. 

To complement the existing hardware, we propose the implementation of a deep learning (DL) computer vision model to examine the images of the pillars in order to classify the damage status. Within this paper, a DL algorithms is trained to classify the presence of damages in VIG images with high accuracy. This work hence furnishes as basis for future research, where training could include learning of the fracture process or cause, once respective production data is fed into the training pipeline. It is then possible that the DL will predict the nature of the origin of the cone crack, identifying issues in quality of the pillar production and/or issues related to inline production handling/processing of the VIG assembly. The current state of the art for visual damage detection is focused on using DL methods, such as convolutional neural networks (CNNs). In the following sections we outline the DL approach, and more importantly, a new variation of a residual neural network that can reduce the time and effort in training an DL algorithm. 

\subsection{Automated Damage Evaluation}
Cracks are highly visible in glass due to the light reflected off the fracture surfaces, and thus, can be examined directly through photographic images. In general, images are complex to analyze due to their lack of data structure \citep{CV}. Especially with images, deep neural networks have been shown to outperform the previous state of the art models \citep{voulodimos2018deep}. The underlying idea of deep neural networks (DNNs) is connecting a large number of small equations in the form of artificial neurons to obtain a complex and adaptable model. The artificial neurons are structured in multiple layers where only the state of the first and the last layer is known \citep{LeCun2015}. All the hidden layers in between are the reason for the denomination \textit{deep} in comparison to the umbrella term of artificial neural networks (ANN). Over time, a variety of different layers have evolved with different functions such as filtering, normalizing or resizing \citep{goodfellow2016deep}.

A common approach for analyzing images are convolutional neural networks (CNN) originally proposed by \citet{726791}. CNN is a class of DNN and is capable of filtering unstructured data in multiple layers for different features at several scales and usually use a large number of small (e.g. 3$\times$3~px) filters. The filter size of a convolutional layer determines the receptive field of its neurons, since they are not connected to every neuron of the previous layers, but only to the neurons within their filter. In the first layer, the filtered features are simple patterns such as straight lines or specific color combinations. The complexity of these features increases with each additional convolutional layer, as each layer can build on the results of the previous ones. With this method, objects and patterns of different sizes can be detected with mostly the same small sized filters, depending on their combination \citep{LeCun2015}.

The different methods for image analysis can roughly be categorised into three different tasks: \textit{classification}, \textit{localisation} and \textit{segmentation}. \textit{Classification} has the goal of classifying an image into predefined categories. The Residual Neural Network by \citet{he2015deep} is one of the best known classification architectures, as it enabled much deeper networks through a skip connection. The skip connection allows data to skip parts of the network. This effectively combines the advantages of a shallow neural network with those of a deep neural network. \textit{Localisation} aims to find, for example, the most accurate box that fits around a searched object. The region-based CNN by \citet{Girshick_2014_CVPR} was the first architecture combining region proposals with CNNs creating a new class of object detection models called R-CNNs. Image \textit{segmentation} is about classifying each pixel of an image. The U-Net by \citet{ronneberger2015unet} was developed for biomedical image segmentation, where particularly little training data is available. The network consists of a downsampling path and an upsampling path connected by skip connections. This allows the model to both recognize details and understand their context. There are also mixed models like the Mask-RCNN by \citet{he2018mask}, combining the abilities of object detection and segmentation models. In general, DNNs are used in almost every part of society, due to their complexity and adaptability \citep{voulodimos2018deep}. Even in structural glass engineering and processing, there are practical applications for such DL models \citep{tubiblio123284,Kraus2020}.

All of the above methods rely on supervised training, in other words where input data (here: images) are given together with labels (here: indicator if damage is present or not). As we cannot access existing data sets for this specific problem, labels must be created manually for all data. Transfer learning techniques can reduce the training time and improve the results and thus achieve good results even with little data \citep{pan2009survey}. However, as we want to optimise a network without compromises for the use of class activation mapping (CAM) methods \citep{zhou2015learning,Selvaraju_2017_ICCV,wang2020scorecam}, we will not use transfer learning.

Depending on how complex the result of a model should be, the more complex the labeling of the data will be. Without a pre-trained model, it is even more important to use the existing data as effectively as possible. In this paper we will focus on how to maximise the results with a smart sequence of actions relying upon explainable AI (XAI) methods without having to train a complex model. This allows us to quickly explore the potential of the data at hand, to estimate the generalization ability more accurately, and to avoid time-consuming mistakes. Despite the large need for XAI, it is not yet widely used due to its technical challenges \citep{8466590}. To that end, this work also serves as a strong advocate and pilot study of XAI methods in civil engineering.

\section{Methods}
\subsection{Data Acquisition}
Data for the image classification task during quality control of the VIGs were generated by photographing the pillars using a microscope. The examined pillars have a diameter of about 0.5 mm. The microscope uses a 1,000-fold magnification lens and an imaging sensor with a resolution of 1,920x1,080 pixels. The brightness is automatically adjusted using control software that monitors the perceived light level.

\begin{figure}[t]
	\centering
	\includegraphics[width=\linewidth]{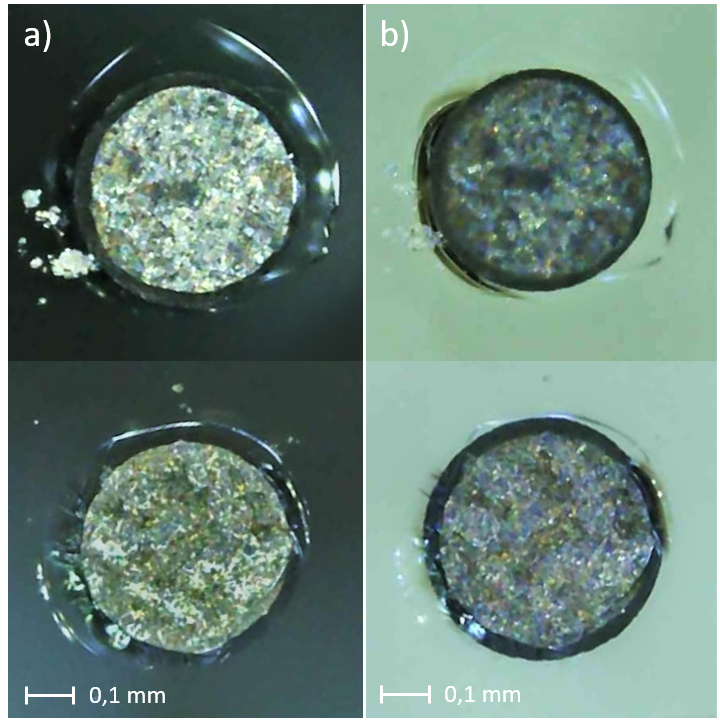}
	\caption{Comparison of the influence of the background color: a) dark background b) light background.}
	\label{fig:Comparison}
\end{figure}

To make the cracks as visible as possible, images taken against different backgrounds, black or white, were compared. Fig.~\ref{fig:Comparison} shows a comparison of two sets of pillar images with a bright and dark background. In fig.~\ref{fig:Comparison}a, the cracks and scratches can be seen more clearly and in greater detail when compared to fig.~\ref{fig:Comparison}b. Due to the bright light reflections at the crack tips and scratches, a dark background is found to be more suitable. However, cracks do not always reflect the light back into the camera, and therefore, may appear dark. Dark cracks can be observed more easily with a bright background. Because "dark cracks" occurred less frequently in our observations, we chose to use the dark background for our investigations in the remainder of this paper.

\begin{figure}[t]
	\centering
	\includegraphics[width=\linewidth]{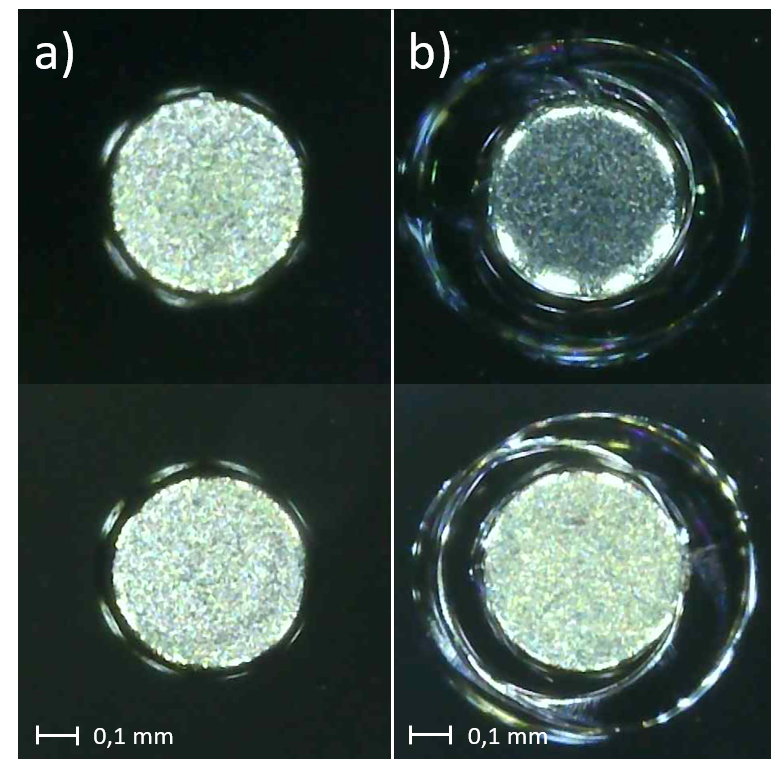}
	\caption{Example pictures for the two classes of pillars: a) undamaged b) damaged.}
	\label{fig:Green and Red Category}
\end{figure}

For the image acquisition, preprocessing and sorting, all steps were performed manually. The microscope was manually positioned and triggered. The images were manually preprocessed and labeled "undamaged" (class 0) and "damaged" (class 1). Since there are no objective criteria for labeling, the categorisation was made according to subjective judgement by human experts. Hence, for the training of the AI algorithm, only images that most clearly belong to the respective category were used. To give an example, fig.~\ref{fig:Green and Red Category}a shows two pillars without damage and fig.~\ref{fig:Green and Red Category}b shows two pillars with damage.

\subsection{Image Preprocessing}
Image preprocessing is crucial to improve the results of a model by processing the training data. For this purpose, we optimized image cropping and resolution, as well as investigated possible augmentation of the images.

\textbf{Resolution Optimization:}  
The input size of a CNN, here the resolution of an image, directly effects the computing time and memory required. It can also affect the number of layers required. If an image has a higher resolution for the same field of view, the objects in it will be represented by a larger number of pixels. In order for a filter of the same size to capture the object at a higher resolution, more downsampling steps are required. Therefore the complexity of the model and the computing time would increase. To avoid these problems, the training data should always be reduced to the relevant information from the beginning. 

A large reduction in data, without loss of information, can be achieved by cropping the images. To ensure that no information is lost, the crop dimension must be optimised to contain the largest expected fracture zone surrounding a pillar. As mentioned previously, pillar contact on glass is simply a process of surface indentation, which is equivalent to a flat rigid indenter on a compliant, nominally flat, elastic half-space \citep{Mouginot1985,kocer1998angle}. The resulting Hertzian stress field is highly tensile at the outer edge of the contact surface and is a diminishing field moving away in all directions from the contact zone. It has been shown that glass fracture due to indentation is described well using the strain energy release rate function (SERR) \citep{Mouginot1985,kocer1998angle}. More importantly, it is found that cone cracks in soda-lime silicate glass typically initiate within a well-defined radial surface zone that is about 1.1 to 1.5 the pillar radius.

Images in this work are, therefore, cropped to an image size that is about 2 times the diameter of the pillar contact, since it is the damage on the contact surface that is of greatest interest. This reduces the imaging resources needed during analysis, while ensuring that the fracture detail of interest is captured in the subsequent image processing steps. In addition, since the pillar is always in different places in the images, initially an algorithm is used to locate the centre of the pillar. We use the Pytorch \citep{paszke2017automatic} implementation of the Faster-RCNN model, originally proposed by \citet{ren2015faster} with a ResNet-50-FPN backbone by \citet{he2015deep}. The ResNet is used to create feature maps of the input image. Those feature maps are used by the model to propose regions of interest. The regions of interest are the attention mechanism of the model and used to classify and locate objects. To create the localization labels of the training data, in this case bounding boxes, the Visual Geometry Group (VGG) annotator called VGG Image Annotator (VIA) was used \citep{dutta2016via}. 300 pictures were annotated and used for training. Further discussion of the localisation model is omitted within this paper as as the main focus of the present paper lays on the damage classification. We refer to the paper from \citet{ren2015faster} for further information regarding the localisation model.

The trained object detection model returns a bounding box per pillar. From the size of the bounding box, the size of the pillar can be estimated, as well as the area in which cracks are expected. This step allows for zooming into the image without losing important information. In the present case, it was possible to reduce the resolution from \(1,000 \times 1,000\)~px to \(700 \times 700\)~px resulting in a pixel count reduction by a factor of two.

To avoid having to process a large image all at once, it can make sense to divide it into small partial images. Dividing the image into arbitrary smaller sections would only be feasible  with a segmentation model, as the label of an image here does not only have a value per image, but a value per pixel. This way we would not have to pay attention to whether there is a crack in each partial image. For our classification model, we can still divide the images into smaller areas by utilising their symmetry. Even though the cracks are not symmetrical, in images with damage, the cracks appear in all quadrants. The prediction is, therefore, the same for each quadrant of an image and symmetrical about the X- and Y-axes. This allows for dividing the image into four parts, while keeping the same labels, quadrupling the size of the dataset.

\begin{figure*}[ht]
	\centering
	\includegraphics[width=\textwidth]{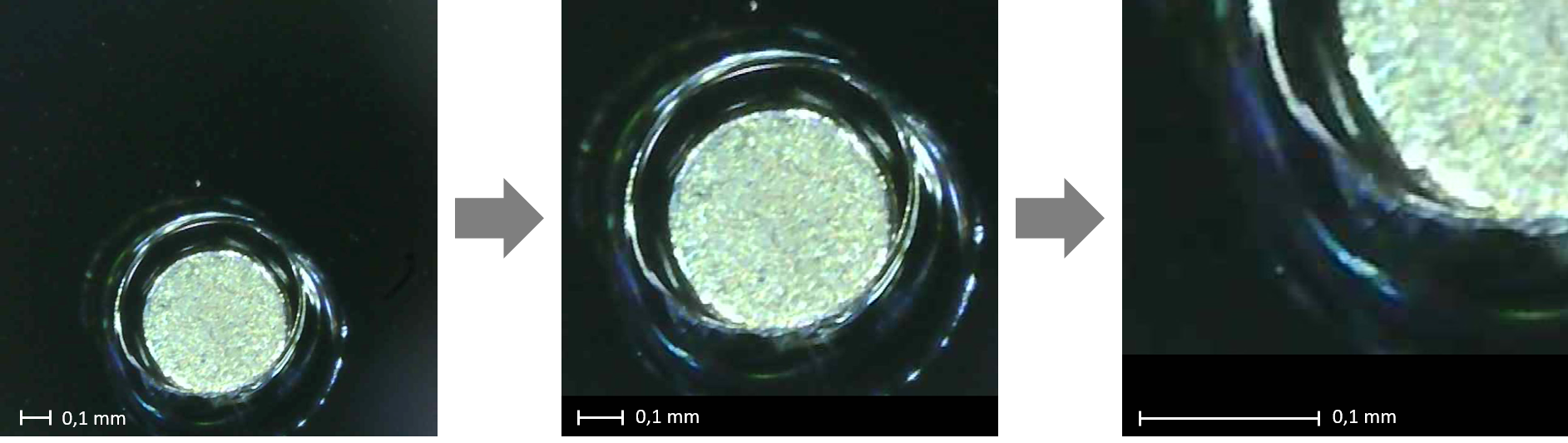}
    \caption{Image preprocessing steps with a) the original image, b) the cropped image based on object detection and c) the quarter image. Zero padding was used to ensure a consistent resolution with a centred pillar.}
    \label{fig:resOpti}
\end{figure*}

Our classification and localisation model are internally halving the resolution several times to detect details of different sizes. To allow the images to be divided without a remainder and to make the partial images overlap slightly, \(352 \times 352\)~px instead of \(350 \times 350\)~px was chosen as the resolution for the partial images. The result is an additional input size reduction by roughly a factor of 4. The three steps are illustrated in Fig.~\ref{fig:resOpti} starting with the raw image on the left, the cropped image in the middle, and the final partial image on the right. As an example, an image with the pillar close to the edge was used. Because the images must have the same resolution and the pillar is must be centred, the remaining area is filled with black pixels, also known as zero padding. Overall, the input size is reduced by a factor of eight compared to the raw data.

\textbf{Augmentation:}  
The main goal in training a model is to achieve good generalisation ability. That is, the ability of the model to make good predictions for new data. Overfitting describes the process where a model fits too closely to the training data, and therefore, does not produce good results with new test data. To achieve a better generalisation ability and prevent the AI algorithm from overfitting, established data augmentation techniques from computer vision are used \citep{shorten2019survey}. Data augmentation effectively increases the amount of training data by altering the training data randomly and operates on the training data set, while other techniques such as dropout \citep{srivastava2014dropout}, batch normalization \citep{ioffe2015batch} or transfer learning \citep{shao2014transfer,weiss2016survey} consider the model itself.

Most importantly, the labels must continue to be correct even if the data was augmented. With a classifier, the labels remain the same even after augmentation, whereas with localisation and segmentation models the labels must also be transformed so that, for example, if the image is mirrored the coordinates of the bounding box must also be mirrored.

\begin{figure}
	\centering
	\includegraphics[width=\linewidth]{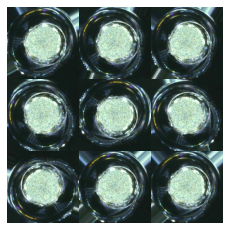}
	\caption{Randomly rotating the image between 0° and 90° with fill mode 'nearest'. Unfavourable patterns occur in the edges of the image.}
	\label{fig:rotation}
\end{figure}

To augment the images, TensorFlows "ImageDataGenerator" \citep{tensorflow2015-whitepaper} was used. We used random rotation with a rotation range of only 10° to ensure that not too much of the image is cut off. Even with small rotation, the augmented images look different each time. Since the image size must remain the same, and the image should not be magnified, the corners must be filled. Because methods such as repeating the nearest value can result in unfavourable patterns (fig.~\ref{fig:rotation}) only zero-filling is feasible in our case. 
Random channel shift was used with a range of 25. The channels in question are, for example, red, green and blue, as is the case with most coloured images. That is, the random channel shift changes the brightness of each channel, or here each colour, separately. Even if the majority of the image is in grayscale, cracks sometimes reflect the light in color. We want to use the information of the color image and still make sure that the model does not react exclusively to colored pixels and, thus, overfit. By randomly changing the brightness of the colors, the information is not lost, but the model cannot rely solely on colored pixels either.

\begin{figure}
	\centering
	\includegraphics[width=\linewidth]{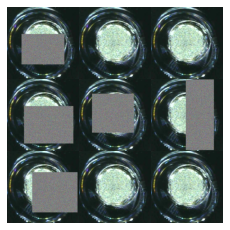}
	\caption{Custom script overwrites 25-50\% of an image with random values for each pixel by a chance of 50\%.}
	\label{fig:erasing}
\end{figure}

In addition, we have used horizontal flip, vertical flip and random brightness ranges from 70\% to 130\%. The random brightness ensures that the model not only classifies the images according to their average brightness, but also detects the fractures themselves. We also extended the built-in functions with a custom script for randomly deleting parts of the image \citep{tubiblio123284}. In our approach, 25-40\% of the image is overwritten with random pixel colour values by a 50\% chance. In fig.~\ref{fig:erasing}, the same image was processed 9 times with this script. Since each pixel of the deleted area is assigned a random color, the deleted area looks gray from a distance. 

\subsection{Classification Model}
Classification is the task of assigning an input to the correct target class. Multi-class neural nets represent each class with one output node. In our case, a single output node is sufficient since both of our classes are mutually exclusive and can thus be represented by a single value. For example, if the model predicts a value of 0.8, this can be interpreted as 80\% probability for class 1 (damaged) and 20\% probability for class 0 (undamaged).

There is a variety of well-known classification models applicable to a wide variety of tasks even beyond computer vision tasks \citep{khan2020survey}. For example, localisation models like the Faster-RCNN \citep{ren2015faster} or segmentation models like the Mask-RCNN \citep{he2018mask} contain a classification model as a so-called backbone. For this project, we developed our own model called VIG damage classification network (VDCNet) based on the residual neural network (ResNet) architecture originally proposed by \citet{he2015deep} and optimized it specifically for the present problem. To evaluate the performance of VDCNet, we compare it with the second generation of ResNets \citep{he2016identity}.

\begin{figure*}
    \centering
    \includegraphics[width=\linewidth]{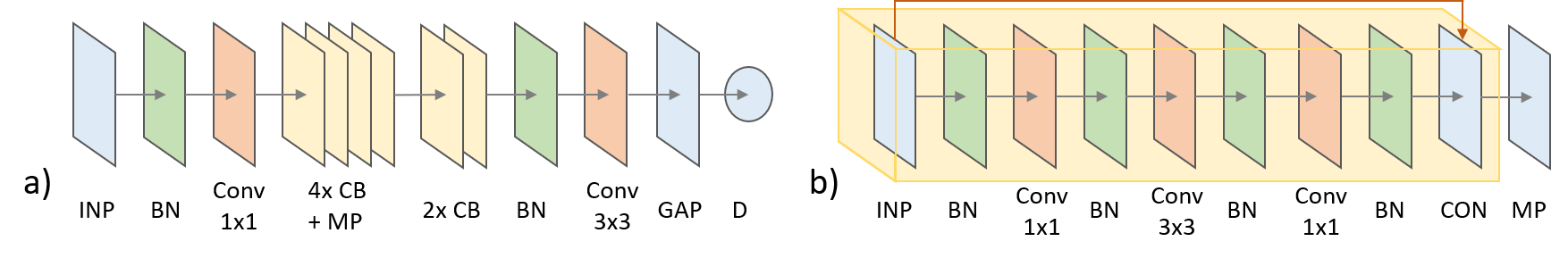}
    \caption{VDCNet architecture with a) the general structure and b) the convolutional block (CB) with a max pooling layer. Arrows symbolize connections between layers, a orange arrow visualizes the skip connection implemented with a concatenate layer, and the other layers: Input image (INP), concatenate (CON), batch norm (BN), 2D convolution (Conv), max pooling 2D (MP), 2D global average pooling (GAP) and dense (D). The figure on the right of the layers indicates the number of their outputs.}
    \label{fig:archi1}
\end{figure*}

The ResNet model features so-called skip connections. In a classical CNN without skip connections, there is often only a single path. This means that errors in one layer are automatically propagated to the other layers and can even be amplified. Instead, in a ResNet, in each filtering block, the input data is passed to two different layers: once to the first filtering layer and once directly to the output layer of the block, which is also known as the skip connection. This output layer in the original ResNet performs an element-wise addition of the filtered data and the data from the skip connection, while the VDCNet concatenates them. Therefore, each layer of the VDCNet receives both the filtered and unfiltered data of the previous layers. This keeps the individual results from each filtering block until a final convolution layer. The drawback is an increasing size of the feature maps with every convolutional block. In order to better understand the architecture shown in fig.~\ref{fig:archi1}, the following section discusses the individual layers of the VDCNet.

\textbf{Convolution Layer:}  
The most important component of a convolutional neural network (CNN) is its name-giving convolution layer. While in a dense layer, all artificial neurons are connected to all input data, a convolution layer is only ever connected to the data in its field of view. The viewing area is then moved over the entire data set for each layer. In the process, the data set, or in our case the image, is filtered each time. In the first layers, small details are filtered, which are combined into larger and more complex features with each subsequent layer. The data filtered by a convolution layer are called feature maps. Thus, a hierarchical structure is used, according to which an object can be seen as the combination of its details \citep{geron}.

The filter size, also known as kernel, can vary depending on the model. Often 3$\times$3~px filters are used because they do not require much memory and can still filter for patterns. For example, if an image is filtered with a 3$\times$3~px identity matrix, mainly oblique lines would stand out. A filter of size 1$\times$1~px cannot filter for patterns on its own, but it can combine previous feature maps, and thus, increase the complexity of the recognised patterns. The weight tensor of such a layer would look like the following: 1$\times$1$\times$\textit{number of previous property maps}$\times$\textit{number of filters} \citep{khan2020survey}. 

Since a 1$\times$1 filter requires 9 times less memory than a 3$\times$3 filter, we have used both kernel sizes in the VDCNet. Similar to the larger ResNet architectures, the kernel sizes of the VDCNet's convolution layers per convolutional block are in the following order: 1$\times$1, 3$\times$3, and 1$\times$1.

\textbf{Pooling Layer:}  
The pooling layer is used to reduce the size of the image. This allows small details to be combined and larger details and patterns to be detected even though the same kernel size is used. The so-called max pooling only passes on the largest value within a pool, depending on its size. With a pool size of 2, the largest value in an area of 2$\times$2~px is passed on. The result is an image with a quarter of the original resolution.

In the reduced image, each pixel represents 4 pixels of the original image. If a filter is applied again, e.g. with a 3$\times$3 kernel, 9 pixels are filtered again per step, but now each pixel represents 4 pixels, i.e. 36 pixels. This cannot be equated with a theoretical filter of the size 6$\times$6 (=36), because even with the smallest pool of 2$\times$2~px, 75\% of the information is lost. However, the filtered properties can be combined in a larger context, whereby greater details can be recognised and the collected information is compressed in a targeted manner \citep{geron}.

The VDCNet has 4 max pooling layers each with a kernel size of 2$\times$2~px to be able to put the details in sufficient context so that it can distinguish between crack and pillar.

\textbf{Batch Norm:}  
The VDCNet has a batch norm (BN) layer before each convolution layer. BN is a technique that was proposed by \citet{ioffe2015batch} to deal with vanishing and exploding gradient problems. It normalizes and zero-centers each input by calculating the mean and standard deviation of the current batch. To ensure that the batch norm also achieves a good result during inference, it uses a moving average and a moving variance calculated with the batches it has seen during training (e.g. in the Keras implementation \citep{chollet2015keras}). 

Using a BN layer as the first layer can make preprocessing like adjusting brightness obsolete \citep{geron}. This can significantly increase performance, especially at an early stage of development without fully optimized image preprocessing. Studies have shown that adding BN layers to a model can reduce the amount of training steps by a factor of 14 to achieve the same accuracy \citep{ioffe2015batch}.

\textbf{Proposed Model Architecture:}  
The model architecture is based on the ResNet architecture and adapted to the specific VIG classification problem at hand. Since the images examined usually show only pillars, cracks, and sometimes debris, the model does not need be as deep as for example the ResNet50V2 model. 

The number of necessary downsampling steps depends on the size of the largest details. In the presented case, 48 pixels seem to be sufficient to distinguish between pillar and crack. Therefore, our model only needs 4 downsampling steps to be able to evaluate details with an original size of 48 pixels with a 3$\times$3 filter since every pixel of the 3$\times$3 filter represents \(2^4\) pixels. We use a single convolutional block per downsampling step and two convolutional blocks after the last downsampling step. 

In total, we employ only 20 convolution layers compared to the 49 convolution layers of ResNet50V2. As a result, the VDCNet has about 25.9 million trainable parameters, a similar number to the ResNet50V2 with 23.5 million trainable parameters, with the difference that the VDCNet has twice the output size in the last convolution layer. This gives us a higher resolution for class activation mapping (CAM) method.

\textbf{Training parameters:}  
Our dataset consists of 322 images, half with and half without damage. These images were divided into quadrants and 21 of these quadrants were manually discarded as they could not be clearly assigned to either class. Thus, 1,267 image quadrants were available, of which 90\% were used for training and 10\% were used for testing. A maximum of 100 epochs were trained with a batch size of 6 and the following callbacks based on validation loss:
\begin{enumerate}
    \item early stopping with a patience of 20 epochs
    \item reduce learning rate on plateau with a patience of 6 by a reduction factor of 0.1
\end{enumerate}
To save computing time, early program termination is used to stop the training if further improvement of the model is not observed. This approach interrupts the model training after a given number of epochs without loss of improvement. This is important since the performance of a model rarely improves strictly monotonically, but most of the time fluctuates exhibiting an improving trend \citep{Prechelt1998}.

By reducing the learning rate when there is a plateau in improvement, we prevent the model from missing its minimum. We had good results with a factor greater than 2 from patience of early stopping to patience of the learning rate reduction. Using this ratio, the learning rate can be reduced two times before training is terminated. It should be ensured that the model is trained as optimally as possible. Therefore, a sufficient number of epochs without improvement must be ensured.

In order to evaluate the performance of our model, we compared it to the second version ResNet models \citep{he2016identity} from the TensorFlow library. To ensure that the influence of random initialization and random train-test-split is excluded, we used a "Stratified K-Folds cross-validator" \citep{scikit-learn} with the same folds for each model. This method divides the training data into 5 folds, maintaining the distribution of classes in each fold. The models are trained 5 times each, with 4 folds as training data and the remaining fold for validation. Each fold is used once for validation so that each combination is used once.

\subsection{Weakly-supervised Object Localization and Explainable Artificial Intelligence Investigations}
Explainable AI methods come in especially handy for technical applications to build trust in AI methods and allow for human perception of crucial patterns. In order to elucidate the internal mechanisms of the developed neural network with respect to its specific decision making, this section is concerned with post-hoc visual explanation methods based on class activation mapping.

Class activation mapping (CAM) originally proposed by \citet{zhou2015learning} enables the use of classification models for localisation tasks without being explicitly trained for it. Since the model is only trained on image-level labels instead of explicit object coordinates, the method can be characterized as weakly-supervised object localization. CAM methods can be used for debugging, sanity checking or the localization itself. 

The original CAM method basically maps the activation value of the last convolution layer for the target class before these results are used for the classification. It can be visualized with a colormap to understand, what the model is looking at. Since the original CAM method, more visual explanations methods, like Gradient-weighted CAM (Grad-CAM) proposed by \citet{Selvaraju_2017_ICCV} or Score-weighted CAM (Score-CAM) proposed by \citet{wang2020scorecam}, were developed. Grad-CAM utilizes the gradients of the specified class from the input until the last convolution layer while Score-CAM calculates the forward score of the target class for each activation map and weights them accordingly. The size of the last convolution layer is decisive for the size of the class activation map. Because the VDCNet has one pooling step less than a ResNet, its class activation map also has 4 times higher resolution.

\section{Results and Discussion}
To find out whether a global average pooling (GAP) layer or a global max pooling (GMP) layer is more suitable as last layer before our dense neuron, we trained both versions under the same conditions. Since the classification performance did not differ significantly, we compared their Grad-CAM results. In this study the Grad-CAM was applied to images of two damaged pillars and two undamaged pillars. Result visualization is then conducted via the OpenCV colormap "plasma" \citep{opencv_library}.

\begin{figure}
    \centering
    \includegraphics[width=\linewidth]{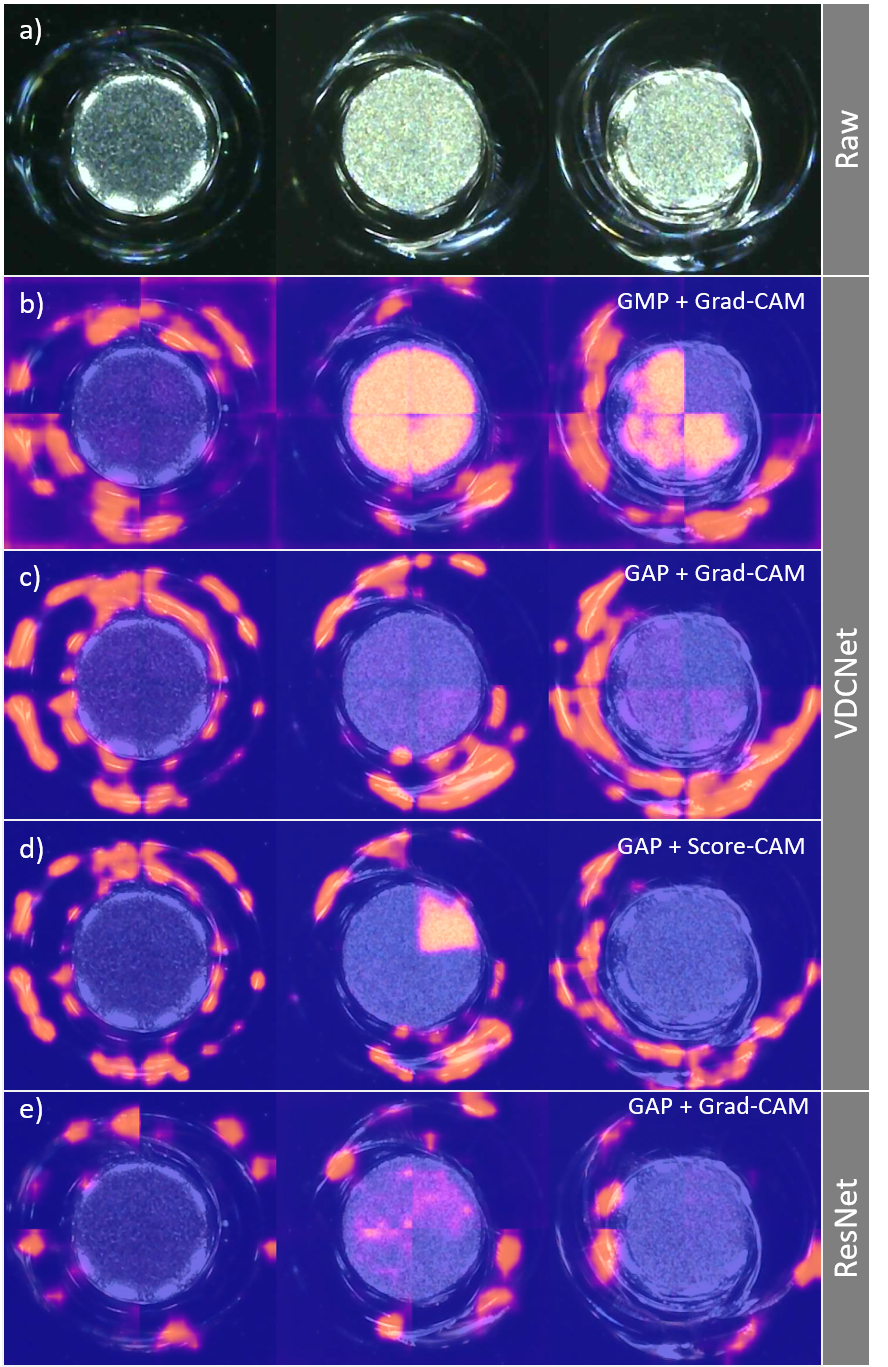}
    \caption{Results of different CAM methods for the same pillars comparing GMP with GAP, Grad-Cam with Score-CAM and VDCNet with ResNet.}
    \label{fig:compare_cam}
\end{figure}

Fig.~\ref{fig:compare_cam}a shows the raw images, fig.~\ref{fig:compare_cam}b shows the results of VDCNet with a GMP layer and fig.~\ref{fig:compare_cam}c shows the results from the same model with a GAP layer. Using a GAP layer, the background is almost completely not activated (dark), while the results with GMP layer are more often activated (brighter), even in places where there is obviously no crack. This is because the GAP layer calculates an average of all activations and thus optimises the model to detect both damaged and non-damaged areas as well as possible. With a GMP layer, only the maximum value of the activation is passed on. The model is thus optimised to find at least one defective spot or, in the case of a non-defective image, the individual values only have to remain below a certain threshold. Because the GMP layer gives worse results than a GAP layer, we choose to only use GAP layers in the model.

Fig.~\ref{fig:compare_cam}d shows the Score-CAM results for the model used in fig.~\ref{fig:compare_cam}c. Most noticeably, the pillar itself is almost not recognized as a crack except for a single quadrant of the pillar in the middle. With the exception of this quadrant, the results seem to be somewhat clearer, but since the calculation time was 30 times longer compared to Grad-CAM (3 hours instead of 6 minutes for 1,288 quadrant images), Score-CAM is therefore out of scope for the present work. The high computation time disqualifies Score-CAM for real-time applications in production lines, although the overall effort may be reduced by using a more efficient implementation in future studies. 

Fig.~\ref{fig:compare_cam}e shows the Grad-CAM results for the \\
ResNet50V2 model. Only spots of the cracks are detected. This already suggests that the model probably lacks a good generalization capability and needs to be revised. It was expected that the VDCNet results would be better suited for localizing the cracks, since its last convolutional layer has twice the resolution compared to ResNet50V2. However, as the ResNet results only detect the cracks selectively, it is the quality of the Grad CAM results and not the lower resolution that is decisive.

\begin{figure}[t]
    \centering
    \includegraphics[width=\linewidth]{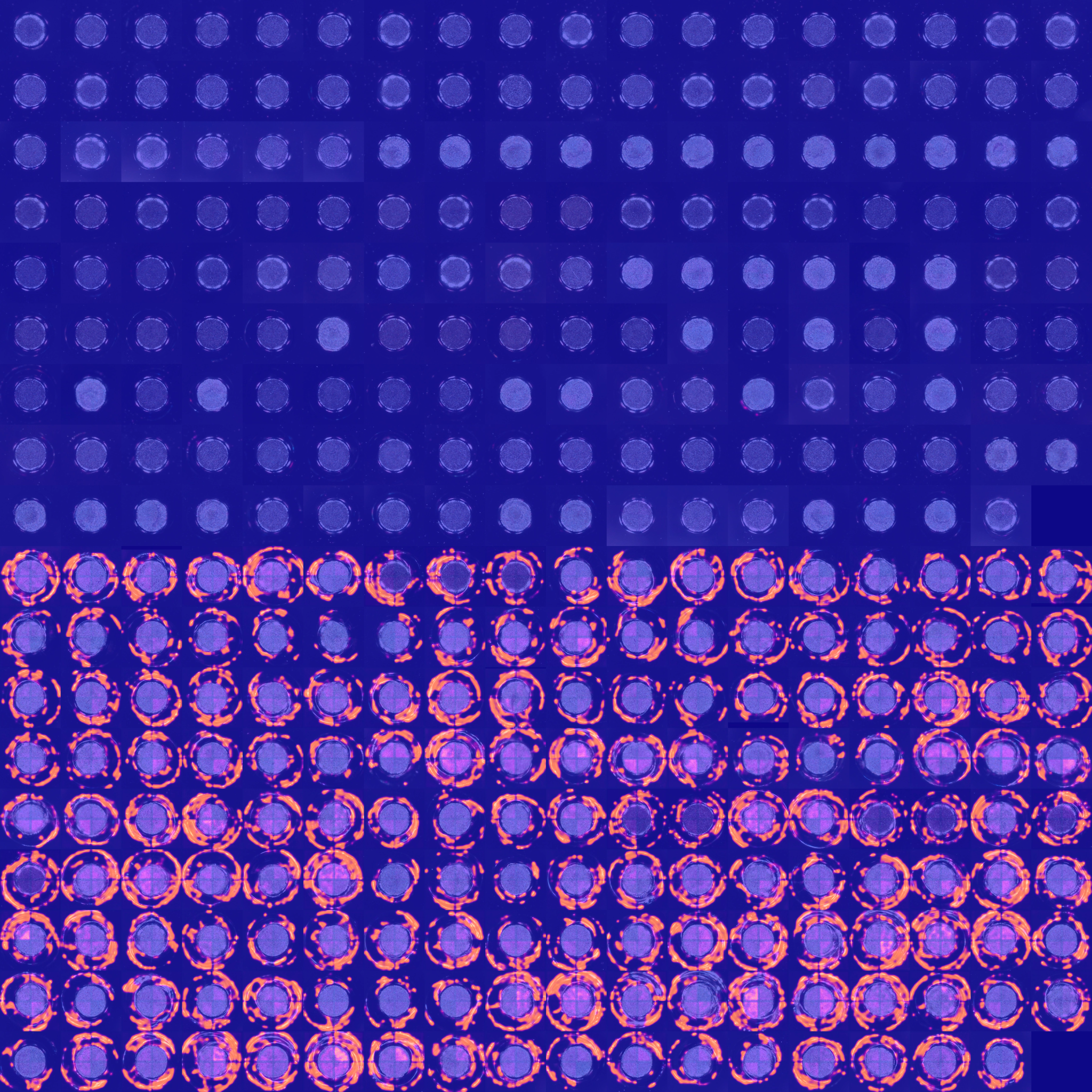}
    \caption{Grad-CAM results of the VDCNet for the training data. The whole dataset of 161 images per category were used with undamaged pillars in the top half and damaged pillars in the bottom half.}
    \label{fig:cam}
\end{figure}

Fig.~\ref{fig:cam} shows the Grad-CAM results of the VDCNet for all 322 pillars. The cracks were not always fully detected, but no clear misclassifications are observed. From this we can conclude that the model detects the cracks themselves and, thus, should give good results even with new data.

\begin{table*}
\centering
\begin{tabular}{  | l | l | l | l | l | l | l | l | l | l | }
\hline
	& \multirow{2}{1cm}{Mean epochs} & \multirow{2}{1.7cm}{Min AUC for ROC} & \multicolumn{2}{c|}{FN 95\%} & \multicolumn{2}{c|}{ FP 10\% } & \multirow{2}{2.5cm}{Mean precision at 100\% recall} & \multirow{2}{1.5cm}{Min accuracy} & \multirow{2}{1.5cm}{Mean loss}  \\ \cline{4-7}
	 & & & mean & max & mean & max & & & \\ \hline
	ResNet152V2 & 69.8 & 0.9981 & 10.8 & 15 & 8.4 & 15 & 0.9475 & 0.9605 & 0.0684 \\ 
	\hline
	ResNet101V2 & 75.6 & 0.9927 & 11.8 & 19 & 20 & 4 & 0.8917 & 0.9842 & 0.0653 \\ 
	\hline
	ResNet50V2 & 67.6 & 0.9989 & 9 & 11 & 13 & 5 & 0.9502 & 0.9881 & 0.0392 \\ 
	\hline
	VDCNet & 49.0 & 1.0000 & 2.4 & 4 & 0.4 & 2 & 1.0000 & 0.9960 & 0.0052 \\ 
	\hline
\end{tabular}
\caption{Summary of VDCNet and ResNet results using the test set.}
\label{table:1}
\end{table*}

For the comparison of the VDCNet with the ResNets, we restored the best weights of the 5 cross-validation runs for each model in each case based on the validation loss and then tested them on the test data. The results for the test data are summarised in the table~\ref{table:1}. As metrics we used the following: maximum amount of training epochs, minimum area under the curve (AUC) for the receiver operating characteristic (ROC), false negatives (FN) with 95\% threshold, false positives (FP) with 10 \% threshold, minimum precision with a threshold selected for 100\% recall, minimum accuracy and mean loss (binary crossentropy). We did not use the F1 score, which is otherwise frequently used, because in our case FN classifications are to be particularly avoided. Therefore, we have set 100\% recall as a boundary condition for the classification threshold and want to know the corresponding precision of the model. Under this constraint, the F1 score is equal to the "precision at 100\% recall".

The VDCNet outperforms the other models in every metric within fewer training epochs. It is the only model to reach consistently 100\% AUC for ROC and 100\% precision with recall at 100\%. VDCNet also had some misclassifications under the stricter conditions of FN with 95\% threshold and FP with 10\% threshold, but they were 4.39 times less frequent for FN classifications and 34.5 times less frequent for FP classifications compared to the ResNets. Under this threshold, VDCNet had a maximum of 4 FN classifications in 127 examples. However, the examples are quadrants and even if damage in all 4 quadrants should not be detected for a pillar, this pillar is only one of hundreds in a VIG. This means that even under these harsh constraints, the risk of an undetected defective VIG is low. Most important is the difference between the models in performance based on precision at 100\% recall. This metric chooses automatically a threshold so that (in our case) no FN classifications occur. With this automatic threshold VDCNet has 100\% precision and no FP classifications. With ResNets, however, the precision only reaches 89-95\% on average, making FP classifications unavoidable even with the test set.

In summary, to apply VDCNet in production quality control lines, given the results, time and labour costs can be significantly improved. On average, positioning the camera and taking, sorting, and documenting the photo for a single pillar takes between 1 to 2 minutes per side. For a VIG with 400 pillars, a total of 800 images must be captured. This corresponds to a time expenditure of up to 1,600 minutes. Some of the investigated VIGs have up to 923 pillars, which leads to a total quality control period, for one VIG unit, of up to 60 hours if conducted manually. In order to reduce the huge processing time and the personal costs, an automated hardware system to manipulate the camera is crucial for industrial deployment of the approach outlined in this paper. A two-way sliding rail as shown in \citet{scratchdetec} could fully automate the image acquisition process. The inspection speed and thus the cost-effectiveness of the proposed quality control approach could be further increased by using an object detection model to selectively capture images of pillars instead of scanning the entire VIG. Further risk analysis could be carried out to reduce the number of pillars or even the number of VIGs to be inspected in order to ensure the desired quality objectives. 

\section{Conclusion}
The glass panes of VIGs are exposed to atmospheric pressure over their service lifetime because of the internal vacuum gap. The pillars between the glass panes, which maintain this gap, can damage the glass under various conditions. Due to the high number of pillars used in practical VIG designs, an automatic damage detection system is highly desirable. In this study, a deep learning method based on the ResNet architecture was used to develop the VDCNet architecture. 

Using explainable AI methods in the form of Grad-CAM and Score-CAM, we confirmed that VDCNet detects surface cracks and not only changes in brightness. It also showed clear advantages over ResNet50V2, due to its higher CAM resolution and more precise activations. Through stratified k-fold cross-validation we were able to confirm that VDCNet has a better generalisation ability in comparison to the ResNet models. VDCNet is the only model that can exclude FP classifications for the test set and requires 27-35\% fewer epochs for training.

The CAM methods allowed us to check the performance of the VDCNet at an early stage of development. We were able to see that it could perform better than a ResNet. If this was not the case, CAM methods would have allowed us to find and debug problems with our model at an earlier stage. Using Grad-CAM, we were able to localise and visualise the detected damage already in a pure classification model. This paper hence is of high value for the glass engineering domain, as it allows for a domain specific interpretation and judging of the relevant image areas to double-check damage identification mechanisms.

For future research, it is interesting to note, that in almost all cases the damage visualised on the glass surfaces at the pillars is found to be caused by macroscopic forces that stem from handling (i.e. transport or installation) or environmental loading (i.e. wind or thermal loads) during service life. In future investigations, the VDCNet could also be directly linked to well-defined mechanical models \citet{kocer2003} (e.g. the fracture mechanics of glass failure at pillar contacts). In linking the models, the imaging process could provide valuable information that would allow damage mitigation, which is a powerful pre- and post-production tool for quality control and ultimately safety in service of the VIG unit. In combination with an automated camera the imaging process will be faster and information about orientation as well as position of the camera then is also available.

\section*{Acknowledgement}
Funding: We gratefully acknowledge financial support by the german Federal Ministry for Economic Affairs and Energy [grant number 03EGB0021B].

\bibliographystyle{spbasic}      
-------------------------------------------------------

\bibliography{AI_VIG}

\end{document}